\documentclass[10pt,twocolumn,letterpaper]{article}

\usepackage{iccv}
\usepackage{times}
\usepackage{epsfig}
\usepackage{graphicx}
\usepackage{amsmath}
\usepackage{amssymb}
\usepackage{booktabs}
\usepackage{xcolor}
\usepackage{soul}
\usepackage{enumitem}

\definecolor{mygreen}{RGB}{50, 160, 60}
\newcommand{\method}{GFM}
\usepackage[pagebackref=true,breaklinks=true,letterpaper=true,colorlinks,bookmarks=false]{hyperref}

\iccvfinalcopy 

\def\iccvPaperID{10512} 

\ificcvfinal\fi

\begin{document}

\title{Towards Geospatial Foundation Models via Continual Pretraining}
\author{Mat\'ias Mendieta$^{1}$\thanks{Work done as an intern at Amazon Web Services}
\quad
Boran Han$^{2}$
\quad
Xingjian Shi$^{3}$
\quad
Yi Zhu$^{3}$ 
\quad
Chen Chen$^{1}$\\
$^{1}$ Center for Research in Computer Vision, University of Central Florida\\
$^{2}$ Amazon Web Services \quad $^{3}$ Boson AI\\
{\tt\small matias.mendieta@ucf.edu \quad boranhan@amazon.com \quad xshiab@connect.ust.hk} \\{\tt\small yi@boson.ai \quad chen.chen@crcv.ucf.edu}
}

\maketitle
\ificcvfinal\thispagestyle{empty}\fi

\begin{abstract}

Geospatial technologies are becoming increasingly essential in our world for a wide range of applications, including agriculture, urban planning, and disaster response. To help improve the applicability and performance of deep learning models on these geospatial tasks, various works have begun investigating foundation models for this domain. Researchers have explored two prominent approaches for introducing such models in geospatial applications, but both have drawbacks in terms of limited performance benefit or prohibitive training cost. Therefore, in this work, we propose a novel paradigm for building highly effective geospatial foundation models with minimal resource cost and carbon impact. We first construct a compact yet diverse dataset from multiple sources to promote feature diversity, which we term GeoPile. Then, we investigate the potential of continual pretraining from large-scale ImageNet-22k models and propose a multi-objective continual pretraining paradigm, which leverages the strong representations of ImageNet while simultaneously providing the freedom to learn valuable in-domain features. Our approach outperforms previous state-of-the-art geospatial pretraining methods in an extensive evaluation on seven downstream datasets covering various tasks such as change detection, classification, multi-label classification, semantic segmentation, and super-resolution. Code is available at \url{https://github.com/mmendiet/GFM}


\end{abstract}
\vspace{-0.5cm}
\section{Introduction} \label{sec:intro}





\begin{figure}
    \centering
    \includegraphics[trim={0 0 0 0},clip,width=0.98\columnwidth]{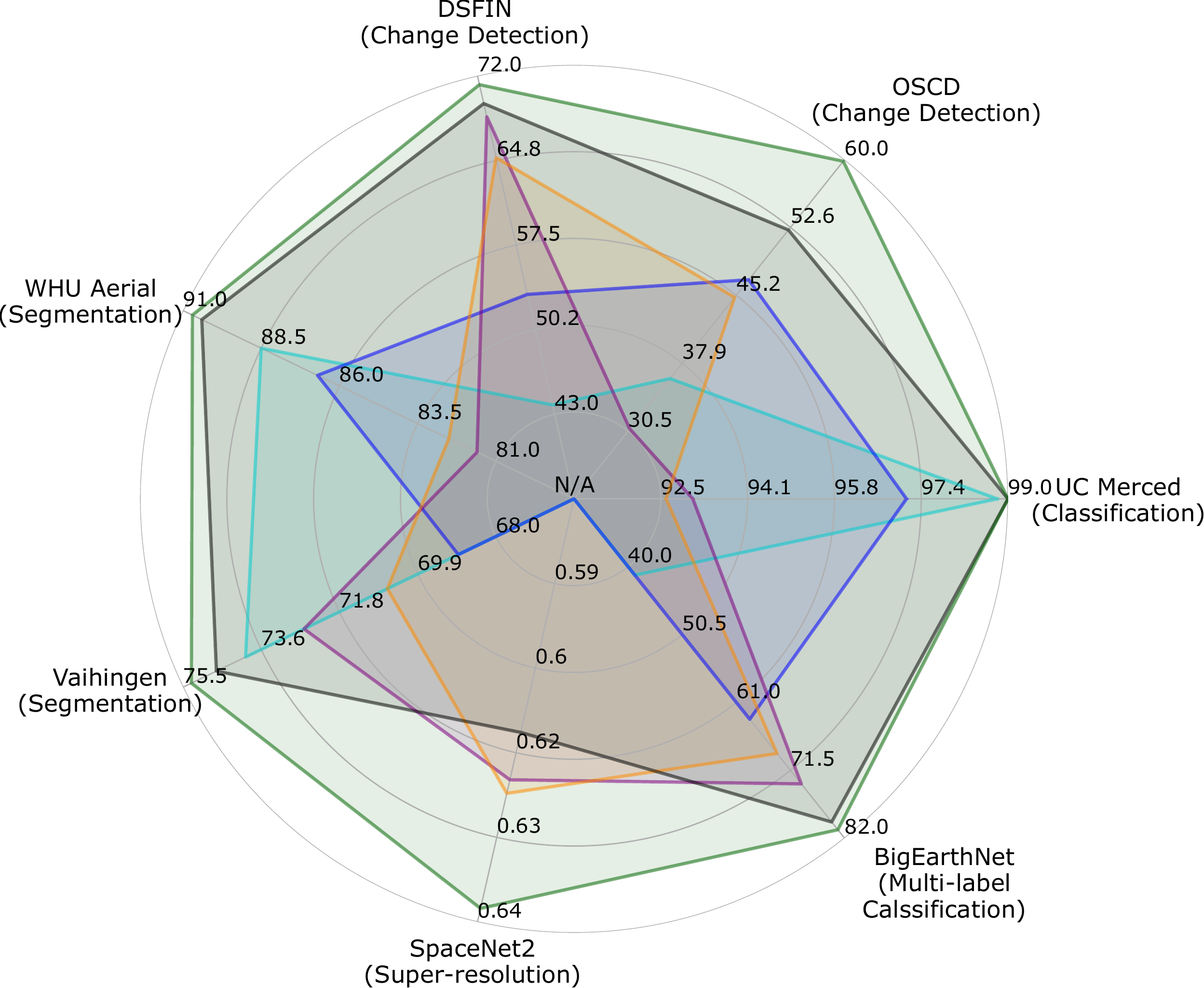}    
    \caption[]
    {Our geospatial foundation model (\method) achieves favorable performance on a broad set of tasks in comparison to other state-of-the-art geospatial pretraining methods (SeCo \cite{seco}, SatMAE \cite{satmae}) and ImageNet supervised pretraining baselines. Legend is as follows. \textcolor{cyan}{Cyan}: ImageNet-1k Supervised (ResNet50), \textcolor{blue}{Blue}: SeCo \cite{seco}, \textcolor{purple}{Purple}: ImageNet-22k Supervised (ViT), \textcolor{orange}{Orange}: SatMAE \cite{satmae}, \textcolor{darkgray}{Gray}: ImageNet-22k Supervised (Swin), \textcolor{mygreen}{Green}: \method~(ours).} 
    \label{fig:model_comparison}
\end{figure}

The significance of geospatial technologies has progressively increased for various applications worldwide.
Progress in this domain can substantially improve our ability to understand the earth and how we interact with it.
With the rising popularity of foundation models in vision and natural language, researchers have begun to investigate applying such principles to the geospatial domain in order to enhance the suitability of deep learning models in downstream tasks \cite{indomain, seco, satmae, gassl}.
In the literature, various works have explored two prominent approaches for introducing pretrained foundation models in geospatial applications. The first obvious approach is to leverage existing foundation models from the natural image domain, like those trained on the large-scale ImageNet-22k dataset \cite{imagenet}. In practice, this is done by \textit{directly finetuning publicly-available ImageNet pretrained models on the downstream tasks}.
This approach has the advantage of being straight-forward, as ImageNet models can be simply downloaded from many open-source model zoos, and has been shown to be effective \cite{indomain, finetune2017}. However, due to the domain gap between natural images and remote sensing, this approach is not optimal for geospatial data, and still leaves performance gains on the table.

In recent years, a second approach has gained significant traction, where researchers aim to pretrained models specific to the geospatial domain \cite{seco, gassl, satmae, millionAID_supervised_pretraining}.
These methods typically \textit{train a network from scratch on a large corpus of remote sensing imagery} to learn in-domain representations transferable to downstream tasks. Unfortunately, this can require a significant amount of data and training time to achieve good performance, especially when employing large state-of-the-art (SOTA) transformer models. For instance, the current SOTA in geospatial foundation models, SatMAE \cite{satmae}, requires 768 hours on a V100 GPU for training a vision transformer \cite{vit}. This has substantial cost associated with producing the model, not just in terms of time and computation but also environmentally, with a total estimated carbon footprint of 109.44 kg CO$_2$ equivalent.
Additionally, the final performance of such models are not consistently better across various tasks than simply utilizing publicly-available ImageNet pretrained models (Section \ref{sec:experiments}),
despite the high resource expense.

In this work, we propose to investigate a different paradigm for producing more effective geospatial foundation models with substantially less resource costs. 
First, we begin with a discussion on pretraining data selection, and ultimately construct a concise yet diverse collection of data from various sources to promote feature diversity and effective pretraining.
Second, rather than following the aforementioned typical approaches, we investigate the potential of \textit{\textbf{continual pretraining for the geospatial domain}} from readily-available ImageNet models.
Continual pretraining has been practiced in the NLP domain with success in various works \cite{dontStopPretraining, continual_temporal, continual_mixedLang}. In this paradigm, existing foundation models are further improved for a specific domain or task through a secondary pretraining stage. This new single model can now be fine-tuned on the various downstream tasks in that domain. In principle, we reason that such a paradigm has the potential to boost performance by utilizing large-scale ImageNet representations as a base on which stronger geospatial foundation models can be built. Furthermore, such natural image models are constantly being improved and released by the general computer vision community, providing a consistent source of better baseline models. Therefore, an approach that could enable the geospatial domain to leverage these improvements with minimal resource needs and carbon footprint paves the way for continual, sustainable benefits for the geospatial community.


However, when we initially experiment with the standard continual pretraining formulation, we find it provides only marginal benefits (Section \ref{sec:continual}). 
Instead, we discover that utilizing ImageNet representations as an auxiliary distillation objective during pretraining leads to a stronger geospatial foundation model. Building upon this principle, we propose a multi-objective continual pretraining paradigm that significantly enhances performance while requiring minimal resources.
Our approach leverages ImageNet's powerful representations to facilitate and expedite learning, while also enabling the acquisition of valuable in-domain features via self-supervised learning on geospatial data. Furthermore, our proposed Geospatial Foundation Model (GFM) exhibits strong performance, surpassing previous state-of-the-art (SOTA) methods across a diverse range of downstream tasks (Section \ref{sec:experiments}).
Our contributions are as follows:
\begin{itemize}
    \item We investigate a novel paradigm for creating highly effective geospatial models with minimal resource costs. Our methodology begins with data selection and construction of a compact yet diverse dataset from multiple sources to promote feature diversity and enhance pretraining effectiveness, which we term GeoPile. We further explore the potential of continual pretraining from ImageNet models, but find it is not satisfactory in its standard formulation.
    \item Therefore, to achieve better performance with minimal resource needs, we propose a multi-objective continual pretraining paradigm. Our design is surprisingly simple yet effective, constructed as a teacher-student strategy with both a distillation objective and self-supervised masked image modeling.
    This approach allows \method~to leverage the strong representations of ImageNet to guide and quicken learning, while simultaneously providing the freedom to learn valuable in-domain features. 
    \item We evaluate our \method~approach, as well as several baseline and SOTA methods, on 7 datasets covering important geospatial applications such as change detection, classification, multi-label classification, semantic segmentation, and super-resolution. Overall, our \method~performs favorably over previous methods (as shown in Figure \ref{fig:model_comparison}).
\end{itemize}

\section{Related Work} \label{sec:related}

\textbf{Geospatial Pretraining}.
Various works have experimented with employing supervised or self-supervised pretraining paradigms in the geospatial domain. The classical work of \cite{indomain}, and  more recent paper \cite{millionAID_supervised_pretraining}, investigate supervised pretraining on individual datasets of various sizes. Interestingly, these still often found the ImageNet pretrained models to perform very well, particularly with vision transformers \cite{vit, swin}.
Other works have explored self-supervised learning paradigms for remote sensing, primarily focused on contrastive methods. \cite{seco} and \cite{gassl} employ a MoCo \cite{mocov2} style objective using spatially aligned but temporally different images as the positive pairs. \cite{saumoco} and \cite{tile2vec} also utilize a MoCo-inspired objective, but specify a cropping procedure to generate positives and negatives within and across images. \cite{colorOutofSpace} employs a colorization objective on Sentinel-2 imagery utilizing the various spectral bands. Most recently, SatMAE \cite{satmae} explores the use of masked image modeling to train a large ViT model. This work is similar in some respect to ours, as we also train a transformer model with an MIM objective. However, we find that SatMAE often does not perform better than the off-the-shelf ImageNet-22k pretrained ViT (Section \ref{sec:experiments}). This indicates both the difficulty of building strong geospatial pretrained models from scratch and highlights the potential usefulness of leveraging continual pretraining instead, as we investigate in this work.

\textbf{Masked Image Modeling}.
Masked image modeling (MIM) has been proposed in various forms in recent years, and has recently been found to be particularly effective in the natural image domain, surpassing many contrastive works and being shown to be friendlier to downstream optimization \cite{simmim, mae, ibot, beit, dark_secrets}
In general, the goal is to learn from data in a self-supervised manner by asking the model to generate pixel values for intentionally-withheld regions in an image.
\cite{context_encoders} is an early work with an aim of learning strong visual representations through inpainting masked regions. In \cite{generative_pretrain}, Chen et. al train a large transformer to predict pixels autoregressively. After the introduction of vision transformers (ViT) \cite{vit}, many works continued to improve various MIM variants. \cite{beit} and \cite{ibot} take inspiration from BERT \cite{bert} in natural language processing, and tokenize the image patches with either a pretrained model or jointly trained online tokenizer, with the objective being to reconstruct at a token-level rather than raw pixels. Recently, \cite{simmim} and \cite{mae} show that a masked image modeling task of simply regressing directly on the image pixels is sufficient and effective. In this work, we leverage the framework from \cite{simmim}, as it is compatible with hierarchical transformer architectures \cite{swin}.

In this work, we develop our pretraining objective based on a masked image modeling approach like \cite{simmim, mae}. Exploration of the masked image modeling framework in geospatial applications is still in its early stages, and could help alleviate some concerns with contrastive approaches in this domain. Particularly, the choice of augmentations with contrastive methods can be quite difficult, as common selections such as greyscale, color jitter and others that heavily affect the intensity of the image can instill undesirable invariances \cite{indomain}. On the other hand, MIM objectives like \cite{simmim, mae} rely only on simple spatial augmentations such as flipping and cropping. Furthermore, a common remote sensing application is that of change detection, which requires a model to detect changes in two images from the same location but at different times. In order to still be effective on this task, works that use contrastive approaches on temporal positives introduce various design choices. For instance, SeCo \cite{seco} creates multiple feature subspaces during pretraining, each one invariant to a separate form of augmentation. \cite{matter} also employs temporal positives, but instead chooses the sampling locations for the pretraining data to ensure that image pairs contain primarily natural illumination and viewing angle variant, without major changes such as new urban developments.

\textbf{Continual Pretraining}.
Continual pretraining has been primarily introduced in the natural language domain \cite{dontStopPretraining, continual_temporal, continual_mixedLang}, in order to improve large language models (LLM). \cite{dontStopPretraining} illustrates the viability of two additional stages of pretraining, using in-domain data (domain-adaptive), and then even further using task-specific data (task-adaptive). \cite{continual_temporal} proposes a continual training paradigm for enabling temporal reasoning abilities to pretrained language models. \cite{continual_mixedLang} focus on using continual pretraining to enable mixed language neural machine translation. In the vision domain, \cite{medseg} employs a BYOL \cite{byol} style continual pretraining paradigm for 2D medical image segmentation. \cite{selfimproveself} explores a hierarchical pretraining approach for task adaptation. However, they primarily focus on adapting to a specific downstream task at a time, employing three training stages on top of an existing pretrained model for each task individually. In contrast, we employ one efficient in-domain pretraining setting that can generalize to many downstream tasks, as illustrated in Section \ref{sec:experiments}. 
Furthermore, rather than directly loading the pretrained weights from existing models as initialization, we find instead that leveraging the representations as an auxiliary distillation objective during the pretraining process enables learning stronger representations.

\section{Methodology}
In the following sections, we discuss the pretraining data selection (Sec, \ref{sec:data}), investigate vanilla continual pretraining (Sec. \ref{sec:continual}), and present our \method~method (Sec. \ref{sec:gfm}).



\subsection{Pre-training Data Selection} \label{sec:data}

A particularly common choice of source data among geospatial contrastive pretraining works is Sentinel-2 imagery \cite{seco, matter, colorOutofSpace} due to its large corpus of available data and ease of access.
Therefore, to begin our study, we first gather a pretraining dataset of ~1.3 million Sentinel-2 images using the sampling technique from \cite{seco}. 
After gathering the Sentinel-2 data, we employ it to pretrain a Swin-B \cite{swin} model with the masked image modeling (MIM) objective from \cite{simmim}. 
We then finetune and evaluate this model on a wide variety of downstream datasets to get a broad understanding of its performance potential in many tasks (see Section \ref{sec:experiments} for task details). For a comparison, we finetune the ImageNet-22k pretrained Swin-B from the official Swin Transformer repository \cite{swin} on all downstream tasks as a baseline. In order to compare these models across all tasks, we introduce an average relative performance metric (ARP) in which we take the relative difference on each task with respect to the ImageNet-22k baseline, and then average that difference:
\begin{equation} \label{eg:arp}
    \text{ARP}(M) = \frac{1}{N}\sum_{i=1}^N \frac{\text{score}(M, \text{task}_i) -\text{score}( \text{baseline}, \text{task}_i)}{\text{score}( \text{baseline}, \text{task}_i)}.
\end{equation}
\noindent Here ``baseline'' is the Swin-B model pretrained on ImageNet-22k, as mentioned above. $M$ denotes the model for performance evaluation, and N is the number of tasks. There are $7$ tasks used in Section~\ref{sec:experiments} covering important geospatial applications such as classification, multi-label classification, semantic segmentation, change detection, and super-resolution. The reported ARP value is scaled by 100 to show as a percentage.

\begin{figure}
    \centering
    \includegraphics[trim={390 160 160 160},clip, width=0.47\textwidth]{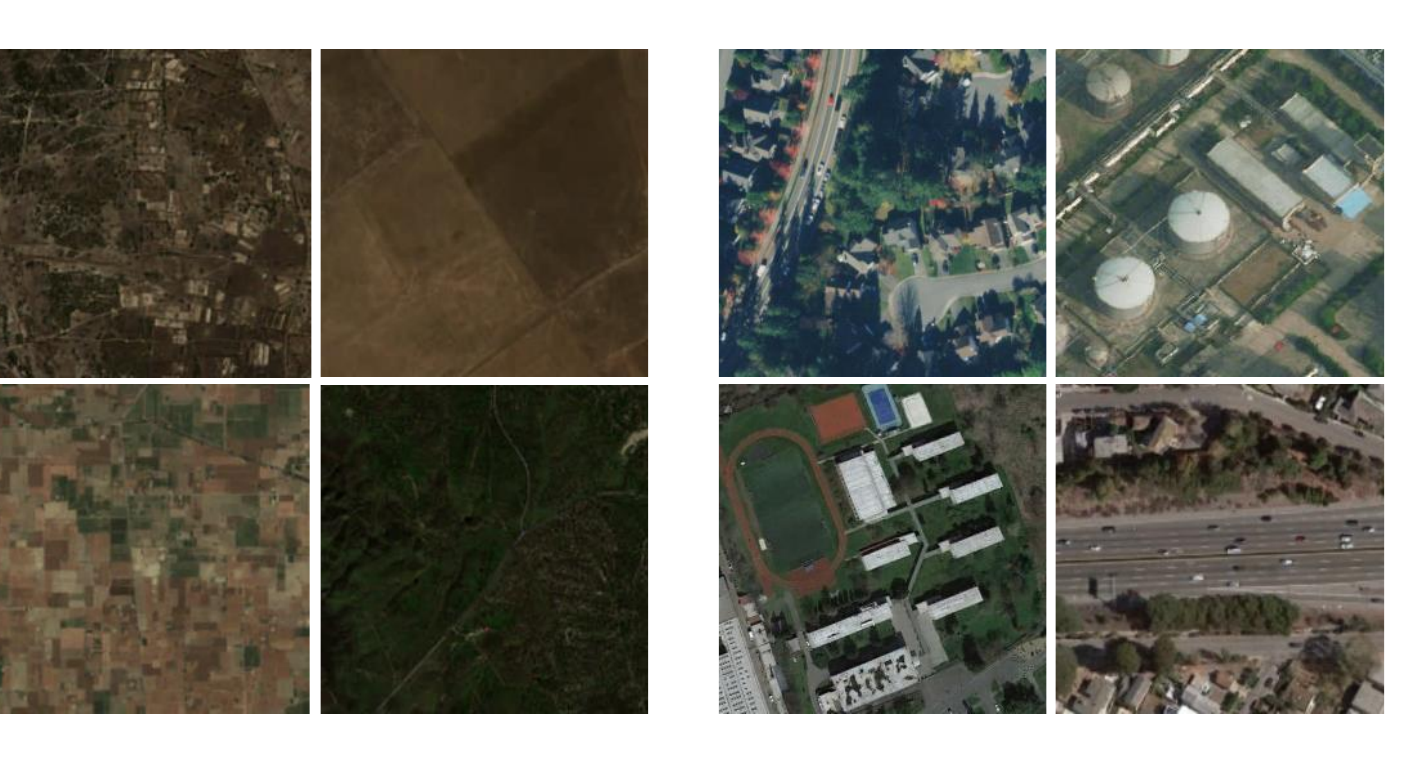}
    \caption[]
    {We visualize some example images from the pretraining datasets with Sentinel-2 (left) and GeoPile (right). Sentinel-2 has noticeably much lower feature diversity within a single image and across images than that of our GeoPile pretraining dataset.}
    \label{fig:data_comparison_visual}
\end{figure}

We compare these two models in Table \ref{tab:data}.
Interestingly, we find that the Sentinel-2 model performs poorly on downstream tasks compared to the ImageNet-22k baseline.
To investigate further, we visualize multiple samples from Sentinel-2 in the left columns of Figure \ref{fig:data_comparison_visual}. Upon inspection, we note that the feature diversity within a single image and across images of Sentinel-2 is perceivably low. To further quantify this suspicion, we calculate the average image entropy over a randomly sampled set of 3000 images from the collected Sentinel-2 data as well as the typical ImageNet dataset as a baseline. Overall, the Sentinel images have an average entropy of 3.9 compared to 5.1 of ImageNet. Such an evaluation provides insights into the potential pitfalls of Sentinel-2 data in pretraining transformers. For MIM objectives, training data with a substantially lower entropy can make for an easier reconstruction task, since masked regions may be more similar to their neighbors. Therefore, the network does not have to work as hard to fill in the blanks, limiting the learning potential.
Overall, these result indicate that the noticeably narrow scope of features and limited per-sample information in Sentinel-2 data may be limiting the potential of the pretrained model.


Therefore, we set out to collect a diverse geospatial pretraining dataset. Sourcing from both labeled and unlabelled data, we form a new pretraining dataset which we term GeoPile. The breakdown of GeoPile is shown in Table \ref{tab:geopile}. For textural detail, we ensure a variety of ground sample distances (GSD), including images with much higher resolution than Sentinel-2 (which has a GSD of 10m). Furthermore, the selected labeled datasets encompass a wide variety of classes from general remote sensing scenes, ensuring visual diversity across samples. We calculate the average entropy of our GeoPile dataset, and find it to be 4.6, much higher than that of Sentinel-2. Furthermore, the textural and visual diversity is qualitatively evident in Figure \ref{fig:data_comparison_visual}. In Table \ref{tab:data}, the enhancing effect of the data selection is clearly shown by the substantial performance increase.

\begin{table}
    \caption{Dataset Analysis. To evaluate each method, we finetune the pretrained model on seven different tasks, outlined in Section \ref{sec:experiments} and report the ARP metric defined in Equation \ref{eg:arp}. We also report the training time in hours on a V100 GPU, as well as the carbon impact estimations\protect\footnotemark~in kg CO$_2$ equivalent \cite{co2}.
    Overall, our collected GeoPile pretraining dataset significantly improves downstream performance. $\dagger$ indicates the vanilla continual pretraining approach of initializing the model with ImageNet-22k weights prior to conducting MIM training on GeoPile. To further improve the performance in an efficient manner, we introduce our continuous pretraining paradigm \method.}
    \label{tab:data}
    \centering
    \setlength\tabcolsep{3.0pt} 
    \small
    \begin{tabular}{cccccc}
        \toprule
        Method & \# Images & Epochs & ARP $\uparrow$ & Time $\downarrow$ & CO$_2$ $\downarrow$\\
        \toprule
        ImageNet-22k Sup. & 14M & - & 0.0 & - & -\\
        \midrule
        Sentinel-2 \cite{seco} & 1.3M & 100 & -5.83 & 155.6 & 22.2\\
        GeoPile & 600k & 200 & 0.92 & 133.3 & 19.0\\
        GeoPile$^\dagger$ & 600k & 200 & 1.24 & 133.3 & 19.0\\
        GeoPile$^\dagger$ & 600k & 800 & 1.45 & 533.2 & 76.0\\
        \midrule
        \method & 600k & 100 & 3.31 & 93.3 & 13.3\\
        \midrule
    \end{tabular}
\end{table}

\footnotetext{CO$_2$ estimations were completed with \href{https://mlco2.github.io/impact}{mlco2.github.io} from \cite{co2}.}

\begin{table}
    \caption{Breakdown of datasets in the GeoPile. We gather approximately 600k samples from a combination of labeled and unlabeled satellite imagery with various ground sample distances and scenes.}
    \label{tab:geopile}
    \centering
    \setlength\tabcolsep{5.0pt} 
    \small
    \begin{tabular}{cccc}
        \toprule
        Dataset & \# Images & GSD & \# Classes\\
        \toprule
        NAIP \cite{naip} & 300,000 & 1m & n/a\\
        RSD46-WHU \cite{RSD46-WHU} &  116,893 & 0.5m - 2m & 46\\
        MLRSNet \cite{MLRSNet} & 109,161 & 0.1m - 10m & 60\\
        RESISC45 \cite{RESISC45} & 31,500 & 0.2m - 30m & 45\\
        PatternNet \cite{PatternNet} & 30,400 & 0.1m - 0.8m & 38\\
        \midrule
    \end{tabular}
\end{table}

\begin{figure*}[t]
    \centering
    \includegraphics[trim={200 175 150 150},clip, width=0.90\textwidth]{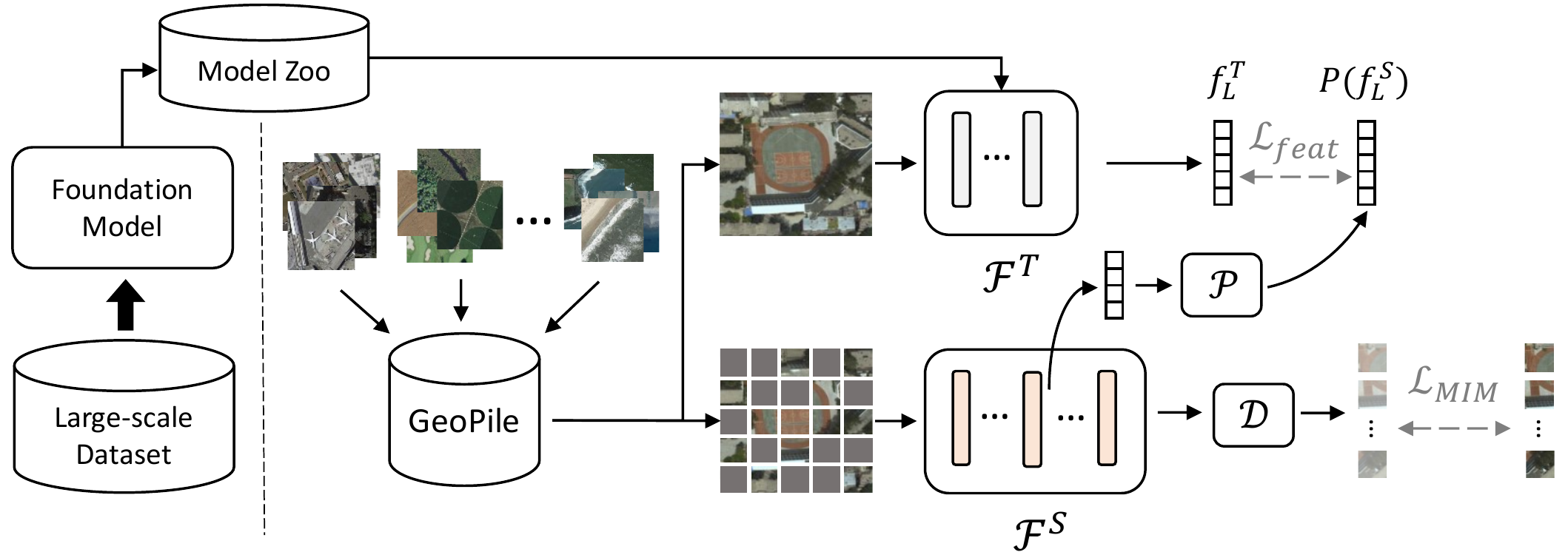}
    \caption[]
    {Our \method~continual pretraining pipeline, which leverages publicly-available large-scale models in concert with our compiled geospatial dataset and pretraining objective. First, we select a concise set of data from various sources, which we term GeoPile (Section \ref{sec:data}). Next, we train \method~with our  multi-objective continual pretraining approach. Our \method~framework is constructed as a teacher-student paradigm, with two parallel model branches. The teacher $\mathcal{F}^{T}$ is initialized with ImageNet-22k weights (top) and frozen during training. The student $\mathcal{F}^{S}$ is initialized from random initialization (bottom), and is trained to serve as the final geospatial foundation model. In a continual pretraining fashion, we leverage the intermediate features of an ImageNet-22k pretrained model to guide and quicken learning. Furthermore, we build in an MIM objective on the student branch to learn valuable in-domain features directly from the geospatial data.} 
    \label{fig:gfm}
\end{figure*}

\subsection{Vanilla Continual Pretraining} \label{sec:continual}
Next, after establishing our pretraining data selection, we investigate an alternate pretraining paradigm that bridges the gap between the two common approaches mentioned in Section \ref{sec:intro}. Specifically, we investigate the potential of continual pretraining in the context of geospatial pretrained models. To do so, we first employ the vanilla continual pretraining approach; that is, using the ImageNet-22k weights as initialization prior to beginning the pretraining step with GeoPile. We find this to be helpful in improving performance over starting from scratch. This validates the possibility of continual pretraining as a beneficial paradigm to provide performance gain without additional resource costs. Nonetheless, the improvement is still limited, with $\sim$0.3\% ARP increase over starting from scratch and $\sim$1.24\% ARP over the baseline.

To further improve the performance of our pretrained model in comparison to the ImageNet-22k baseline, we increase the number of pretraining epochs in the next row of Table \ref{tab:data}. While we are able to make improvements, this comes at the cost of substantially more computational cost and carbon footprint for marginal gain. Therefore, we ask the question: how can we significantly improve the performance further while maintaining minimal compute and carbon footprint overhead? To this end, we propose a simple and efficient approach for building geospatial pretrained models capable of strong downstream performance.
\subsection{\method~Pretraining} \label{sec:gfm}

A significant number of geospatial foundation model studies disregard the existing large-scale model representations. This is far from ideal, particularly for large transformer models known to require a vast amount of data and compute power to train. Instead, we reason that the valuable knowledge available in models like those trained on ImageNet-22k should be leveraged to produce strong performance with minimized overhead.
To this end, we propose an unsupervised multi-objective training paradigm for effective and efficient pretraining of geospatial models, illustrated in Figure \ref{fig:gfm}.

There are two main components in our framework. First, we randomly initialize an encoder $\mathcal{F}^{S}$ and decoder $\mathcal{D}$ set up for MIM as in \cite{simmim}. During training, the input is randomly masked, and the network attempts to reconstruct the image at the output. This MIM objective is enforced with an L1 loss \cite{simmim}:
\begin{equation}
    \mathcal{L}_{MIM}=\frac{\left\|\mathbf{O}_\kappa-\mathbf{G}_\kappa\right\|_1}{N},
\end{equation}
where $\mathbf{O}_{\kappa}$ are the original pixel values from $\kappa$ masked regions, $\mathbf{G}_{\kappa}$ are the generated reconstructions for those regions, and $N$ is the total number of masked pixels.

For the continual pretraining of our framework, we initialize a second encoder branch $\mathcal{F}^{T}$ up to a chosen stage $L$ and load the ImageNet-22k pretrained weights. This branch behaves as a form of teacher during the training process to the student branch ($\mathcal{F}^{S}$), which will serve as our final model. For the ImageNet teacher, we freeze the weights, to both ensure that the structured representations are maintained during the training process, and also reduce the computation required during optimization. 
 
Rather than using the masked input as in the student branch, the teacher receives the unmasked image as input,
and provides a feature output $f_{L}^{T}$ at stage $L$. This feature has access to the full context of the input, enabling it to capture informative representations.
We utilize this feature to guide the representations of the student, and form a secondary objective with the cosine similarity between branch features: 
\begin{equation}
    \mathcal{L}_{feat} =  -\frac{P(f_{L}^{S})}{\left\|P(f_{L}^{S})\right\|_2} \cdot \frac{f_{L}^{T}}{\left\|f_{L}^{T}\right\|_2},
\end{equation}
where $f_{L}^{S}$ and $f_{L}^{T}$ are the intermediate features of the student and teacher branches at stage $L$, and $\mathcal{P}$ is an linear projection layer. Therefore, the final loss during training is simply the summation of these objectives:

\begin{equation} \label{eq:loss_combined}
    \mathcal{L} = \mathcal{L}_{MIM} + \mathcal{L}_{feat}.
\end{equation}
This training paradigm enables an ideal two-fold optimization. Distillation from the intermediate features of the teacher ensure that the student can benefit from the teacher's diverse knowledge, learning more in less time. Furthermore, the student is simultaneously given freedom to adapt to in-domain data through its own pretraining objective, gathering new features to improve performance.

We analyze the ARP and resource cost of this approach in Table \ref{tab:data}. Notably, our \method~is able to achieve better overall performance with substantially less computation and emissions impact compared to vanilla continual pretraining with the same dataset, illustrating that our multi-objective continual pretraining paradigm is an effective method for training these models. 
Comparatively, the SOTA geospatial pretrained method SatMAE \cite{satmae} requires 768 hours on a V100 GPU and 109.44 kg equivalent CO$_2$ according to their reported results. Therefore, \method~enables more than 8× reduction in total training time and carbon impact. Moreover, we find that the performance of SatMAE is often not superior to the off-the-shelf ImageNet-22k pretrained ViT (Section \ref{sec:experiments}). This implies that building powerful geospatial pretrained models from scratch is challenging and further underscores the benefits of utilizing continual pretraining instead. We show these results in the following section.



\section{Experiments} \label{sec:experiments}
To verify the effectiveness of our model in detail, we conduct experiments on seven geospatial datasets of various tasks including change detection (Section \ref{sec:change_det}), classification (Section \ref{sec:classification}), segmentation (Section \ref{sec:seg_detect}), and super-resolution (Section \ref{sec:superres}). 

For pretraining, we employ 8 NVIDIA V100 GPUs with a batch size of 2048 (128 per GPU) and the image size of 192×192. All pretraining settings are the same as in \cite{simmim}. For downstream tasks, 4 NVIDIA A10G GPUs are employed. 
During the pretraining stage, we utilize RGB bands as they are most commonly available among data sources and tasks.  For downstream tasks with additional band inputs, we initialize the RGB patch embeddings with the pretrained weights and randomly initialize the remaining channels. 
Potentially improving performance even further though the employment of additional data modalities will be an intriguing avenue for future research. Additional training details for these tasks are provided in the \textit{supplementary material}.
\begin{table}
    \caption{Onera Satellite Change Detection Results}
    \label{tab:OSCD}
    \centering
    \setlength\tabcolsep{5.0pt} 
    \resizebox{\columnwidth}{!}{
    \begin{tabular}{cccc}
        \toprule
        Method & Precision $\uparrow$ & Recall $\uparrow$ & F1 $\uparrow$\\
        \toprule
        ResNet50 (ImageNet-1k) \cite{resnet} & \textbf{70.42} & 25.12 & 36.20\\
        SeCo \cite{seco} & 65.47 & 38.06 & 46.94\\
        MATTER \cite{matter} & 61.80 & 57.13 & 59.37\\
        ViT (ImageNet-22k) \cite{vit} & 48.34 & 22.52 & 30.73\\
        SatMAE \cite{satmae} & 48.19 & 42.24 & 45.02\\
        Swin (random)\cite{swin} & 51.80 & 47.69 & 49.66\\
        Swin (ImageNet-22k)\cite{swin} & 46.88 & 59.28 & 52.35\\
        \midrule
        \method & 58.07 & \textbf{61.67} & \textbf{59.82}\\
        \midrule
    \end{tabular}
    }
\end{table}

\begin{figure}
    \centering
    \includegraphics[width=0.9\columnwidth]{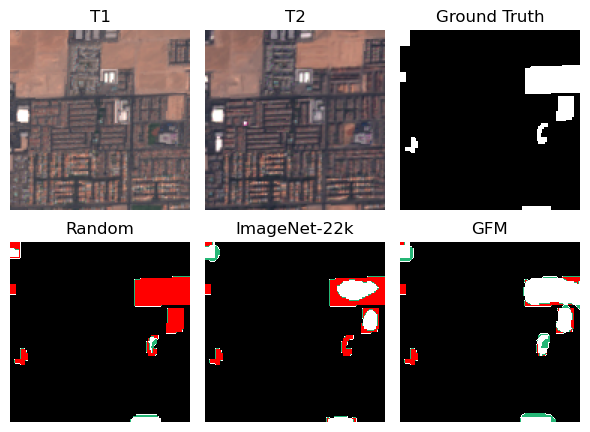}
    \caption[]
    {Qualitative results of downstream performance on OSCD comparing our \method~with ImageNet-22k and randomly initialized baselines. White, green, red colors show true positive, false positive, and false negative respectively.} 
    \label{fig:OSCD}
\end{figure}

\begin{table}
    \caption{DSFIN Change Detection Results}
    \label{tab:DSFIN}
    \centering
    \setlength\tabcolsep{5.0pt} 
    \resizebox{\columnwidth}{!}{
    \begin{tabular}{cccc}
        \toprule
        Method & Precision $\uparrow$ & Recall $\uparrow$ & F1 $\uparrow$\\
        \toprule
        ResNet50 (ImageNet-1k) \cite{resnet} & 28.74 & \textbf{92.07} & 43.80\\
        SeCo \cite{seco} & 39.68 & 81.02 & 53.27\\
        ViT (ImageNet-22k) \cite{vit} & 70.77 & 66.34 & 68.49\\
        SatMAE \cite{satmae} & 70.45 & 60.29 & 64.98\\
        Swin (random)\cite{swin} & 57.97 & 62.06 & 59.94\\
        Swin (ImageNet-22k)\cite{swin} & 67.11 & 72.33 & 69.62\\
        \midrule
        \method & \textbf{74.83} & 67.98 & \textbf{71.24}\\
        \midrule
    \end{tabular}
    }
\end{table}

\subsection{Change Detection} \label{sec:change_det}
Change detection is a particularly important remote sensing task, helping us understand how humans interact with our planet over time, and natural phenomena that change our planet's landscape. We conduct experiments on both the Onera Satellite Change Detection (OSCD \cite{OSCD}) in Table \ref{tab:OSCD} and DSIFN \cite{DSIFN} in Table \ref{tab:DSFIN}.


OSCD consists of 14 image pairs extracted from various regions around the world within a three year period of 2015 to 2018. The images are taken from Sentinel-2 with GSDs ranging from 10m to 60m, and split into 14 images for training and 10 for evaluation. The annotations indicate whether the change has occurred on a pixel level, and focus primarily on urban developments. Similarly, we also test our method on DSIFN dataset. This dataset contains high-resolution imagery, such as WorldView-3 and GeoEys-1 \cite{DSIFN}. This dataset contains 3490 high resolution samples for training and 48 images for evaluation respectively. Every pair of images from a given location at two different timestamps will be fed into the swin encoder \cite{swin} for feature extraction. The difference between the features from each pair is computed and fed into an UPerNet \cite{Upernet} to generate the final binary segmentation masks \cite{seco, siamdiff}. The encoder is initialized with the pretrained weights.


For both datasets, we report the precision, recall, and F1 score on the ``change" class. As the results presented from OSCD (Table \ref{tab:OSCD} and Figure \ref{fig:OSCD}) and DSIFN (Table \ref{tab:DSFIN}), \method~shows a consistent improvement over the ImageNet-22k baseline across both datasets. Notably, SatMAE is able to improve over its ImageNet-22k baseline on OSCD, but lags behind on DSIFN. This further highlights the difficulty of training large vision transformers from scratch that can perform consistently across different GSDs. 

\subsection{Classification} \label{sec:classification}
Another common remote sensing application is that of classification. We evaluate two datasets common in the literature \cite{seco, matter}: UC Merced Land Use Dataset \cite{ucm} and BigEarthNet \cite{BEN}.
The UC Merced Land Use Dataset is a classic dataset in the remote sensing field. It contains 21 classes, each with 100 images at 256x256 pixels and an approximate GSD of 1 foot. We split the data into train and validation according to \cite{data_splits}. BigEarthNet \cite{BEN} (BEN) is a large-scale remote sensing dataset for multi-label classification. The data consist of 12-band Sentinel-2 images with sizes of 120x120, 60x60, and 20x20 pixels for the bands at 10m, 20m, and 60m GSDs, respectively.
We employ the data split and 19 class evaluation as common in the literature \cite{indomain, seco, satmae}.

In Table \ref{tab:BEN}, we report the classification accuracy on UC Merced (UCM) and mean average precision results on BigEarthNet (BEN) for all methods.
On UC Merced, we note the SeCo \cite{seco} pretrained model performs significantly worse than its ImageNet-1k pretrained counterpart with ResNet-50. 
These two datasets are very different in both classes, satellite source, and GSDs, and therefore having a diverse feature knowledge is imperative to maintaining performance despite these distinctions.
Our model can provide robust performance in both cases by leveraging ImageNet representations and remote sensing data in its learning. Furthermore, one key motivation for training a geospatial foundation model is to improve the sample efficiency for downstream tasks. Notably, we find that our model maintains strong performance on BigEarthNet, even when only given 1\% of the training data.

\begin{table}
    \caption{UC Merced classification accuracy and BigEarthNet multi-label classification mean average precision results.}
    \label{tab:BEN}
    \centering
    \setlength\tabcolsep{5.0pt} 
    \resizebox{\columnwidth}{!}{
    \begin{tabular}{cccc}
        \toprule
        Method & UCM  & BEN 10\% & BEN 1\%\\
        \toprule
        ResNet50 (ImageNet-1k) \cite{resnet} & 98.8 & 80.0 & 41.3\\
        SeCo \cite{seco} & 97.1 & 82.6 & 63.6\\
        ViT (ImageNet-22k)\cite{vit} & 93.1 & 84.7 & 73.6\\
        SatMAE \cite{satmae} & 92.6 & 81.8 & 68.9\\
        Swin (random)\cite{swin} & 66.9 & 80.6 & 65.7\\
        Swin (ImageNet-22k) \cite{swin} & \textbf{99.0} & 85.7 & 79.5\\
        \midrule
       \method & \textbf{99.0} & \textbf{86.3} & \textbf{80.7}\\
        \midrule
    \end{tabular}
    }
\end{table}

\subsection{Segmentation} \label{sec:seg_detect}
Segmentation is a popular remote sensing application for enabling automated extraction of building footprints or land cover mappings over wide regions. We therefore conduct experiments on this task on two different datasets.
Vaihingen \cite{vaihingen} is an urban semantic segmentation dataset collected over Vaihingen, Germany at a GSD of 0.9m. We employ the data split implemented in the MMSegmentation library \cite{mmseg} for our experiments, with 344 training and 398 for validation, all with an image size of 512x512 pixels. The WHU Aerial building \cite{whu} dataset is sampled over Christchurch, New Zealand at a GSD of 0.3m. Image tiles are provided at $512\times 512$ pixels, split into 4736 for training and 2416 for evaluation.

We report the intersect of union (IoU) segmentation results for all methods in Table \ref{tab:seg}. ImageNet pretrained models are notably strong performers in all cases. On both datasets, SeCo lags substantially behind its ImageNet counterpart. Interestingly, SatMAE is able to bring improvement over ImageNet-22k on WHU, but fails to do so to a larger degree on Vaihingen. 
However, our approach is able to leverage the already strong ImageNet-22k representations and guide them towards the geospatial domain, resulting in overall improvement.
\begin{table}
    \caption{Results on the WHU Aerial and Vaihingen segmentation datasets. We finetune all methods for 40k iterations, and report the IoU for the building class on WHU and mean IoU (mIoU) across the 6 classes (impervious surface, building, low vegetation, tree, car, clutter) of Vaihingen.}
    \label{tab:seg}
    \centering
    \setlength\tabcolsep{5.0pt} 
    \small
    \begin{tabular}{cccc}
        \toprule
        Method & WHU Aerial & Vaihingen\\
        \toprule
        ResNet50 (ImageNet-1k) \cite{resnet} & 88.5 & 74.0\\
        SeCo \cite{seco} & 86.7 & 68.9\\
        ViT (ImageNet-22k) \cite{vit} & 81.6 & 72.6\\
        SatMAE \cite{satmae} & 82.5 & 70.6 \\
        Swin (random) \cite{swin} & 88.2 & 67.0\\
        Swin (ImageNet-22k) \cite{swin} & 90.4 & 74.7 \\
        \midrule
        \method & \textbf{90.7} & \textbf{75.3} \\
        \midrule
    \end{tabular}
\end{table}

\subsection{Super-resolution} \label{sec:superres}
In the previous experiments, we evaluated several common high-level tasks. Nonetheless, the low-level task of super-resolution is also important in the geospatial domain.
For this task, we re-purpose the SpaceNet2 dataset, which contains 10,593 8-band images from four cities: Las Vegas, Paris, Shanghai, and Khartoum. The data is provided at both a GSD of 1.24m (multi-spectral, 162x162 pixels) and 0.3m (pan-sharpened multispectral, 650x650 pixels). We formulate a super-resolution task, taking as input the 1.24m multi-spectral images and generating the 0.3m pan-sharpened equivalent. We evaluate the super-resolution performance of our model and several baselines with the peak signal-to-noise ratio (PSNR) and structural similarity index measure (SSIM) in Table \ref{tab:spacenet}.
The ViT-L ImageNet-22k model and our model are among the best in terms of PSNR and SSIM, respectively. Interestingly, SatMAE is not able to improve over its baseline. On the other hand, our method improves considerably over its ImageNet-22k baseline.

\begin{table}
    \caption{SpaceNet2 Super-resolution Results. Notably, while SatMAE fails to enhance its baseline (ViT ImageNet-22k), our method exhibits substantial improvement over its respective baseline (Swin ImageNet-22k) in both PSNR and SSIM.}
    \label{tab:spacenet}
    \centering
    \begin{tabular}{ccc}
        \toprule
        Method & PSNR $\uparrow$ & SSIM $\uparrow$\\
        \toprule
        ViT (ImageNet-22k)\cite{vit} & \textbf{23.279} & 0.619 \\
        SatMAE \cite{satmae} & 22.742 & 0.621 \\
        Swin (random) \cite{swin} & 21.825 & 0.594 \\
        Swin (ImageNet-22k) \cite{swin} & 21.655 & 0.612 \\
        \midrule
        \method & 22.599 & \textbf{0.638} \\
        \midrule
    \end{tabular}
\end{table}

\section{Ablation Studies} \label{sec:ablation}

We perform multiple ablation studies on the choice of distillation stage, student initialization, training objectives, the pretraining dataset components. Further detailed results and discussions are provided in the \textit{supplementary material}.

\subsection{Distillation Stage}
\begin{figure}
    \centering
    \includegraphics[width=0.95\columnwidth]{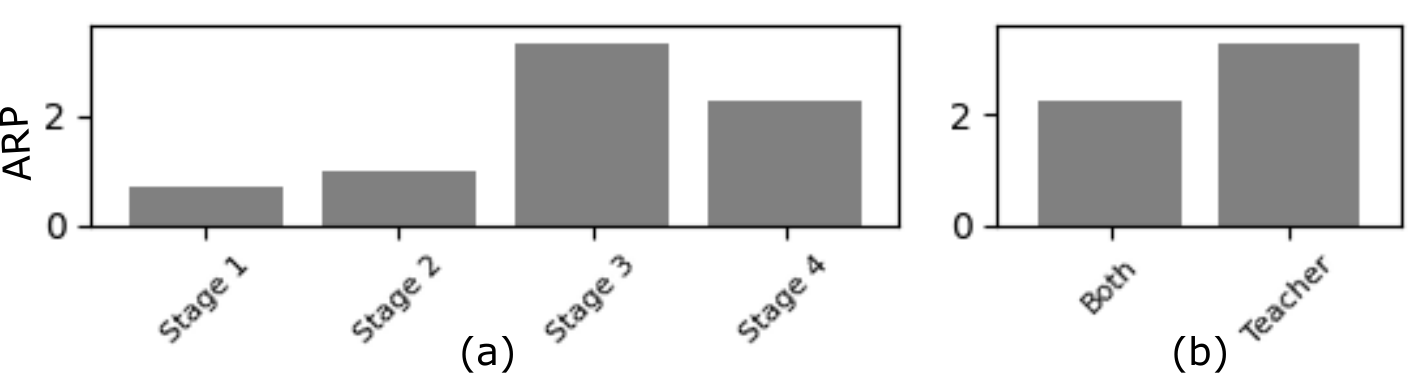}
    \caption[]
    {a) Distillation stage ablation results. b) Student initialization ablation results. ``Both" indicates that the teacher and student branches are initialized with ImageNet weights prior to geospatial pretraining. ``Teacher" indicates that just the teacher branch is initialized, as described in Section \ref{sec:gfm}.}
    \label{fig:ablation_plot}
\end{figure}

When implementing our feature map distillation objective, a natural question is at which point should the mapping take place. We experiment with different locations by stage in the Swin transformer and calculate the corresponding ARP in Figure \ref{fig:ablation_plot}. Overall, performing the distillation after Stage 3 yields the highest ARP. Hence, we employ this scheme for all downstream experiments.
This result is also intuitively expected; distilling at Stage 3 gives a large portion of the model the supervisory signal from the teacher, while still allowing for purely domain-specific feature learning in the final layers.

\subsection{Student Initialization}
In our proposed framework, we maintain the teacher model frozen with ImageNet pretrained weights, and randomly initialize the student. Another alternative is to initialize the student also with ImageNet weights prior to beginning the geospatial pretraining process. However, as shown in Figure \ref{fig:ablation_plot}, this is not the most optimal option. Such initialization is unnecessary in our framework, since it already allows for seamless integration of ImageNet representations with valuable in-domain features. 
Forcibly doing so likely introduces too much bias towards the natural image representations.
Therefore an unbiased student is most ideal and effective. 

\begin{table}
    \caption{GeoPile pretraining dataset ablation. We remove each dataset individually from GeoPile and report the number of images remaining and resulting ARP. The row ``w/o curated datasets" removes all data other than NAIP imagery.}
    \label{tab:data_ablation}
    \centering
    \begin{tabular}{ccc}
        \toprule
        Data & \# Images & ARP $\uparrow$ \\
        \toprule
        w/o WHU-RSD46 & 444,061 & 1.77\\
        w/o MLRSNet & 451,793 & 2.17\\
        w/o Resisc45 & 529,454 & 1.57\\
        w/o PatternNet & 557,554 & 1.79\\
        w/o curated datasets & 300,000 & 0.53\\
        w/o NAIP & 260,954 & 1.50\\
        \midrule
    \end{tabular}
\end{table}
\subsection{GeoPile Pretraining Dataset}
To ablate components of the GeoPile, we remove each dataset individually to see its relative importance. Also, we compare using just the labeled data portion and using just the unlabeled NAIP imagery portion. As expected, using just data from labeled datasets gives better performance with less images than using just images gathered from just NAIP. The human-curated samples in these datasets are more likely to contain relevant objects and features, as they each correspond to a particular class of interest. Still, unlabeled data like NAIP can be sourced easily and with scale. Further scaling of both labeled and unlabeled portions could further improve performance; however, it will also increase the training time and sustainability impact. Therefore, we maintain GeoPile at approximately 600,000 images.
\begin{table*}[h]
    \caption{Ablation results for the training objectives in GFM. For w/o teacher, we only conduct MIM with GeoPile. For w/o MIM, we simply perform the distillation objective from the ImageNet-22k model to our student model with GeoPile. We abbreviate the following for horizontal space: UC Merced (UCM), BigEarthNet (BEN), WHU Aerial (WHU), Vaihingen (Vai), SpaceNet2 (SN2).}
    \label{tab:min_ablation}
    \centering
    \small
    \begin{tabular}{cccccccccc}
        \toprule
        Method & OSCD (F1) & DSFIN (F1) & UCM & BEN 10\% & BEN 1\% & WHU & Vai. & SN2 (PSNR) & SN2 (SSIM)\\
        \toprule
        w/o teacher & 57.3 & 67.65 & 98.8 & \textbf{86.5} & 80.0 & 90.5 & 74.0 & 22.509 & 0.631\\
        w/o MIM & 59.58 & \textbf{71.86} & 98.8 & 86.1 & 80.2 & 90.2 & 72.6 & 22.069 & 0.608\\
        \midrule
        GFM & \textbf{59.82} & 71.24 & \textbf{99.0} & 86.3 & \textbf{80.7} & \textbf{90.7} & \textbf{75.3} & \textbf{22.599} & \textbf{0.638}\\
        \bottomrule
    \end{tabular}
\end{table*}

\begin{table*}[h]
    \caption{Results for employing temporal pairs and datasets from SeCo \cite{seco} in our multi-objective pretraining framework. TP indicates that the teacher receives one image from a temporal pair, and the student receives the other. SI indicates that the same image is inputted to the teacher and student.}
    \label{tab:data_temp}
    \centering
    \small
    \begin{tabular}{ccccccccccc}
        \toprule
        Dataset & Inputs & OSCD (F1) & DSFIN (F1) & UCM & BEN 10\% & BEN 1\% & WHU & Vai. & SN2 (PSNR) & SN2 (SSIM)\\
        \toprule
        SeCo 100k \cite{seco} & TP & 57.03 & 62.48 & 80.0 & 80.6 & 68.6 & 88.3 & 66.3 & 22.078 & 0.572\\
         SeCo 100k \cite{seco} & SI & 58.41 & 67.92 & 92.1 & 83.9 & 76.5 & 88.8 & 68.1 & 22.439 & 0.602\\
        SeCo 1M \cite{seco} & SI & 58.87 & 69.41 & 95.7 & 86.2 & 77.1 & 89.6 & 71.0 & 22.281 & 0.626\\
        \midrule
        GeoPile & SI & \textbf{59.82} & \textbf{71.24} & \textbf{99.0} & \textbf{86.3} & \textbf{80.7} & \textbf{90.7} & \textbf{75.3} & \textbf{22.599} & \textbf{0.638}\\
        \bottomrule
    \end{tabular}
\end{table*}
\subsection{Multi-objective Ablation.} \label{mim_ablation}
To delve deeper into the evaluation of GFM's performance, we extend our analysis by conducting experiments in which we exclude the teacher component and MIM component individually, as detailed in Table \ref{tab:min_ablation}. 
We find that training with the multi-objective approach is the best performer overall. 
This shows that the integrated distillation and MIM objectives within the GFM framework both contribute to producing a well-balanced mode for downstream tasks, and are important aspects of efficient and effective geospatial learning.

\subsection{Temporal Pairs Experiment} \label{temporal}
Some works employ temporal pairs in the pretraining procedure \cite{seco, gassl, matter}, meaning two satellite images from the same spatial region but taken at different times. We also experiment with the use of temporal positives in our training paradigm using the dataset proposed in SeCo \cite{seco}. In this case, the teacher receives one image from a temporal pair, and the student receives the other. The temporal changes can possibly serve as a form of natural augmentation for the distillation objective. However, as shown in Table \ref{tab:data_temp}, we find that using temporal positives (TP) is worse than simply using the same image (SI) for both branches. Therefore, we simply use the same image for both branches for other experiments. We further scale up the data by employing the 1M sample Sentinel-based dataset from SeCo. Nonetheless, GeoPile proves to be more effective as a pretraining data source for our GFM.

\section{Conclusion}
In summary, this paper investigates an alternative paradigm from previous work towards producing better geospatial foundation models with substantially less resource cost. To this end, we first construct a concise yet diverse collection of data from various remote sensing sources for pretraining. Second, we propose a surprisingly simply yet effective multi-objective continual pretraining paradigm, in which we leverage the strong representations of ImageNet-22k to guide and quicken learning, while simultaneously providing the freedom to learn valuable in-domain features through self-supervised learning on geospatial data.
We hope that our \method~approach will serve as an example to inspire other works in investigating efficient and sustainable methods for developing geospatial foundation models.

\textbf{Broader Impact and Limitations.} 
As the geospatial community continues to innovate, the resulting impact promises to positively benefit both the earth and society. Automating the process of extracting useful information from geospatial data can aid scientists, engineers, and others to make data-informed decisions on infrastructure advancement, food supply improvements, and natural disaster response. A potential limitation of our \method~approach is that it may still be somewhat constrained by the performance of the ImageNet-22k model. If perhaps a model was trained from scratch on an extremely large corpus of remote sensing data, the performance may eventually also lead to improved performance over ImageNet baselines. However, this would incur a substantial amount of training time and CO$_2$ impact. Furthermore, as mentioned in Section \ref{sec:intro}, natural image models are constantly being improved and released by the general computer vision community. Therefore, our approach enables the geospatial domain to effectively leverage these improvements for better in-domain performance with minimal carbon impact. We believe this is a sustainable way for the geospatial community to continually benefit from the most recent progress in computer vision, enabling a smarter, safer, and healthier planet. 

{\small
\bibliographystyle{ieee_fullname}
\bibliography{egbib}
}

\clearpage
\appendix
\section*{Supplementary Material}
The supplementary material is organized into the following sections:

\begin{itemize}
    \item Section~\ref{training_details}: Training details for the pretraining stage and all downstream tasks.
    \item Section~\ref{carbon}: Details on calculations of CO$_2$ impact.
    \item Section~\ref{superres_residual}: Further analysis on the SpaceNet2 super-resolution task.
\end{itemize}

\section{Training Details} \label{training_details}
We provide the training details for the various stages and tasks in our evaluation. Code, model weights, and GeoPile dataset are publicly available at \url{https://github.com/mmendiet/GFM}.

\textbf{Change Detection}:
We modify the MMsegmentation \cite{mmseg} framework to conduct our change detection experiments.
For OSCD, as the raw image size is large but the number of samples is very small, we tile the images into 192$\times$192 pixels and train for 4000 iterations. We utilize the RGB bands for OSCD as in \cite{seco}. For DSFIN, we train for 10k iterations with image size 512$\times$512. We employ an SGD optimizer with a learning rate of 0.01 and weight decay of 5.0e-4, and the default polynomial scheduler of \cite{mmseg}.

\textbf{Classification}:
On UC Merced, we train with a batch size of 1024 (128 per GPU) at image size 256$\times$256. We train for 100 epochs with a base learning rate of 1.0e-4. We employ random flip, crop and standard Mixup \cite{mixup} augmentation. Optimizer, weight decay, Mixup parameters, and other training settings are the same as in \cite{simmim}.
For BigEarthNet, we slightly upscale the original 120$\times$120 images to 128$\times$128 for ease of dimensional compatibility with the Swin transformer. We then employ the same training settings as with UC Merced.

\textbf{Segmentation}:
We employ the MMsegmentation \cite{mmseg} framework to conduct our segmentation experiments. For both datasets, we train for 40k iterations with an image size of 512$\times$512. All other training settings are the same as the default configuration in \cite{mmseg} for the respective backbones (Swin, ViT, ResNet50) and compatible decoders (UperNet \cite{Upernet} for transformers and Deeplabv3 \cite{deeplab} for ResNets).

\textbf{Super-resolution}:
On the SpaceNet2 super-resolution tasks, we train with a batch size of 64 (16 per GPU) with input image size 160$\times$160 and target size 640$\times$640. We train for 100 epochs with a base learning rate of 1.25e-5. Optimizer, weight decay, and other training settings are the same as in \cite{simmim}, but with no random augmentations. We employ the standard decoder from \cite{simmim} to produce the original input size from the encoder features, and then upscale using a convolution-based upsampling block based on the image reconstruction module for classic super-resolution employed in \cite{swinir}.
Detailed results for all downstream experiments and ablations from the main manuscript are provided in Table \ref{tab:full_results}.

\section{Training Time and Carbon Calculations} \label{carbon}

To calculate the CO$_2$ impact of training various models, we employ the ML CO$_2$ Impact estimator at \url{https://mlco2.github.io/impact} from \cite{co2}. The total impact is dependent on the hardware type, GPU provider, region, and total time used. Our pretraining experiments were conducted in the AWS US East (Ohio) region, which has a carbon efficiency of 0.57 kg eq. CO2 per kWh. For our GFM, just 93.3 V100 GPU hours are needed for training, resulting in a total carbon impact of 13.3 kg eq. CO2. This is significantly lower than the previous state-of-the-art geospatial model, SatMAE \cite{satmae}. According to the reported carbon impact in their paper \cite{satmae}, SatMAE requires 768 V100 GPU hours and 109.44 kg eq. CO2 on the Google Cloud Platform us-central1 region, which has a carbon efficiency of 0.57 kg eq. CO2 per kWh (same as AWS US East Ohio). Therefore, \ul{GFM enables more than 8$\times$ reduction in total training time and carbon impact in comparison to SatMAE.}




\section{Super-resolution with Residual Connection} \label{superres_residual}
In super-resolution tasks, a residual connection can be included from the input to the output stage \cite{swinir}. We make this modification as well for both ViT and Swin, and present the results in Table \ref{tab:spacenet_residual}. Interestingly, the Swin transformer benefits from this, while ViT does not. Nonetheless, in comparison to baselines, the conclusion is the same; SatMAE is not able to improve over its ImageNet-22k baseline, but GFM does.

\begin{table}
    \caption{SpaceNet2 super-resolution results with the residual connection.}
    \label{tab:spacenet_residual}
    \centering
    \setlength\tabcolsep{5.0pt} 
    \begin{tabular}{ccc}
        \toprule
        Method & PSNR $\uparrow$ & SSIM $\uparrow$\\
        \toprule
        ViT (ImageNet-22k)\cite{vit} & 22.548 & 0.629 \\
        SatMAE \cite{satmae} & 22.450 & 0.636 \\
        Swin (random) \cite{swin} & 22.190 & 0.642 \\
        Swin (ImageNet-22k) \cite{swin} & 22.918 & 0.640 \\
        \midrule
        GFM & \textbf{22.963} & \textbf{0.660} \\
        \midrule
    \end{tabular}
\end{table}

\begin{table*}[t]
    \caption{Detailed downstream results for all experiments in the main manuscript. We abbreviate the following for horizontal space: UC Merced (UCM), BigEarthNet (BEN), WHU Aerial (WHU), Vaihingen (Vai), SpaceNet2 (SN2). $\dagger$ indicates vanilla continual pretraining.}
    \label{tab:full_results}
    \centering
    \setlength\tabcolsep{2.5pt} 
    \begin{tabular}{cccccccccc}
        \toprule
        Method & OSCD (F1) & DSFIN (F1) & UCM & BEN 10\% & BEN 1\% & WHU & Vai. & SN2 (PSNR) & SN2 (SSIM)\\
        \toprule
        ImageNet-22k baseline & 52.35 & 69.62 & 99.0 & 85.7 & 79.5 & 90.4 & 74.7 & 21.655 & 0.612\\
        \midrule
        Sentinel-2 & 55.14 & 64.31 & 94.5 & 84.9 & 70.0 & 86.2 & 63.3 & 19.961 & 0.566\\
        GeoPile & 56.59 & 68.31 & 98.8 & 86.0 & 79.2 & 89.4 & 73.6 & 22.315 & 0.630\\
        GeoPile$^\dagger$ & 57.10 & 66.88 & 98.7 & 86.2 & 79.3 & 90.0 & 74.6 & 22.566 & 0.638\\
        GeoPile$^\dagger$ (800ep) & 57.52 & 66.23 & 98.8 & 86.3 & 79.3 & 90.1 & 75.1 & 22.626 & 0.645\\
        \midrule
        Stage 1 & 56.20 & 69.79 & 98.1 & 85.8 & 78.3 & 89.0 & 73.3 & 22.153 & 0.626\\
        Stage 2 & 58.97 & 68.27 & 96.9 & 86.1 & 79.0 & 89.4 & 72.2 & 22.409 & 0.625\\
        Stage 4 & 60.31 & 68.97 & 98.3 & 86.1 & 80.8 & 89.8 & 73.0 & 22.495 & 0.638\\
        \midrule
        Both Init. & 58.01 & 69.77 & 98.5 & 85.8 & 77.2 & 90.1 & 74.1 & 22.930 & 0.669\\
        \midrule
        w/o WHU-RSD46 & 58.79 & 69.25 & 98.3 & 86.1 & 80.6 & 89.7 & 72.9 & 22.510 & 0.632\\
        w/o MLRSNet & 60.01 & 69.21 & 98.8 & 86.1 & 80.5 & 89.9 & 72.9 & 22.409 & 0.633\\
        w/o Resisc45 & 58.33 & 69.22 & 98.6 & 86.3 & 80.7 & 89.8 & 72.4 & 22.206 & 0.635\\
        w/o PatternNet & 59.00 & 70.37 & 98.3 & 86.3 & 80.5 & 89.8 & 71.9 & 22.293 & 0.629\\
        w/o curated datasets & 58.49 & 67.16 & 98.1 & 85.7 & 79.9 & 88.9 & 72.7 & 22.852 & 0.584\\
        w/o NAIP & 58.72 & 70.54 & 98.3 & 85.5 & 79.6 & 89.7 & 70.8 & 22.574 & 0.632\\
        \midrule
        GFM & 59.82 & 71.24 & 99.0 & 86.3 & 80.7 & 90.7 & 75.3 & 22.599 & 0.638\\
        \bottomrule
    \end{tabular}
\end{table*}
\end{document}